%% file: detector.tex
\documentclass[twocolumn]{article}

\usepackage{amsmath}
\usepackage{color}
\usepackage[top=1.2in,bottom=1.5in,left=15mm,right=15mm]{geometry}
\usepackage{graphicx}
\usepackage{calc}
\usepackage{caption}
\usepackage{mathptmx}
\usepackage{natbib}
\usepackage{subcaption}
\usepackage{textcomp}
\usepackage[normalem]{ulem}
\usepackage[hidelinks]{hyperref}

\providecommand{\abs}[1]{\lvert#1\rvert}
\newcommand{\centring}{\centering}

\begin{document}

\title{ChESS -- Quick and Robust Detection of Chess-board Features}

\date{December 5, 2012}
\author{Stuart Bennett and Joan Lasenby\thanks{S. Bennett and J. Lasenby are with Cambridge University Engineering Department, Trumpington Street, Cambridge, CB2 1PZ, United Kingdom.  email:~\{sb476, jl221\}@cam.ac.uk}}

\maketitle

\begin{abstract}
Localization of chess-board vertices is a common task in computer vision, underpinning many applications, but relatively little work focusses on designing a specific feature detector that is fast, accurate and robust.  In this paper the ``Chess-board Extraction by Subtraction and Summation'' (ChESS) feature detector, designed to exclusively respond to chess-board vertices, is presented.  The method proposed is robust against noise, poor lighting and poor contrast, requires no prior knowledge of the extent of the chess-board pattern, is computationally very efficient, and provides a strength measure of detected features.  Such a detector has significant application both in the key field of camera calibration, as well as in Structured Light 3D reconstruction.  Evidence is presented showing its robustness, accuracy, and efficiency in comparison to other commonly used detectors both under simulation and in experimental 3D reconstruction of flat plate and cylindrical objects.\\[1em]
\end{abstract}

\section{Introduction}

Many applications in machine vision depend on having accurately localized the vertices of a chess-board pattern, since such patterns are commonly used in camera calibration.  The available methods for this process tend to suffer in the face of severe optical distortion and perspective effects, and often require hand-tuning of parameters, depending on lighting and pattern scale.  Manual intervention is time consuming, requiring operator skill and prohibiting automated use.

Another application needing precise vertex detection is 3D surface reconstruction, where a chess-board pattern is employed as part of a simple yet accurate structured light projector-camera system.  For such use feature extraction must be highly automated and fast.

We will present a robust process specifically targeting the detection of chess-board pattern vertices, which rather than requiring a binary vertex/not-vertex threshold, provides a measure of strength similar in output to the much-used \cite{harris1988} corner detector in the same problem-space.  This permits deferral of inclusion decisions to a later stage where one is better able to exploit spatial and geometric considerations.

The process is also computationally efficient and well disposed to a variety of parallel processing techniques, with a reference implementation capable of throughput of over 700 VGA resolution frames per second on commodity PC hardware.

\section{Related work}

There are various published techniques used for finding the intersections in chess-board patterns, typically employed during a camera calibration routine, though relatively little work focusses on an optimal detector for such commonly used features.  As observed in \cite{soh1997}, regarding camera calibration: ``it is often assumed that the detection of such charts or markers which are designed to enhance their detectability is trivial''.  They continue that this assumption is ill-advised, as generic approaches, such as that of \cite{canny1986}, do not make use of the specific properties of the features and are likely to suffer under sub-optimal conditions of lighting and object pose, and furthermore they highlight the risks of employing lossy ``remedies, such as wide kernel filtering, which are notorious for degrading the positional information and shape of critical features''.

Soh et al. go on to describe a grid processing scheme using a chain of Sobel operators, local thresholding, non-maximal suppression, edge joining, geometric constraints and finally taking the centres of gravity of the found squares.  As \cite{escalera2010} noted, such localization methods were abandoned since the centres of gravity do not coincide with the centres of the squares due to perspective effects.  De la Escalera and Armingol also make a similar point to Soh et al., that less attention has been paid to locating the points used in calibration algorithms than to the calibration algorithms themselves -- a deficit this work aims to redress.

De la Escalera and Armingol's detection scheme uses the Harris and Stephens corner detector to locate a grid, before employing the Hough transform on the image to enforce linearity constraints and discard responses from the corner detector which do not lie along strong linear features.  This then restricts the method's use in applications where the grid is potentially distorted, be it due to optical distortion or a non-planar surface.  They discount the use of corner positions alone (in a situation where the grid boundary is unknown) due to the excessive number of non-grid corners likely to be found by Harris and Stephens' general-purpose algorithm elsewhere in a scene.

\cite{yu2006} describe an alternative method of finding features, which attempts to pattern-match a small image of an intersection by measuring the correlation of this pattern over all the captured image.  Unless a number of such small images are tested however, this method is clearly at a disadvantage when the grid is rotated relative to the intersection view stored in the pattern.

\begin{figure}
	\setbox0=\hbox{\includegraphics{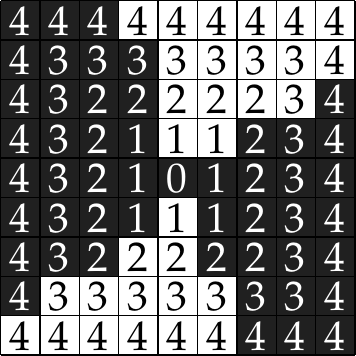}}
	\setbox1=\hbox{\includegraphics{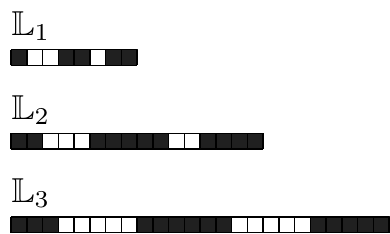}}
	\ifx\halfdiff\undefined
	\newlength{\halfdiff}
	\fi
	\setlength{\halfdiff}{0.5\ht0-0.5\ht1}
	\centring
	\begin{subfigure}{0.49\linewidth}
		\centring
		\unhbox0
		\caption{The rectangular sampling window, with layers numbered}
		\label{fig:sun_a}
	\end{subfigure}
	\begin{subfigure}{0.49\linewidth}
		\centring
		\vspace{\halfdiff}
		\raisebox{\halfdiff}{\unhbox1}
		\caption{The linearized layers 1-3, starting on the top horizontal row and working clockwise}
		\label{fig:sun_b}
	\end{subfigure}
	\caption{Illustration of the \cite{sun2008} sampling window and layer scheme}
	\label{fig:sun_layers}
\end{figure}

Finally \cite{sun2008} detail a method where they pass a rectangular or circular window over the captured image and for each position transform the 2D points distribution along the perimeter of this window into a 1D vector.  For each ring concentric with the perimeter another vector is formed similarly, each vector being termed a `layer', as numbered in \autoref{fig:sun_a} and linearized in \autoref{fig:sun_b}.  The layers are binarized using a locally adaptive threshold, open and close morphological operations applied (\citealt{haralick1987}), and the positions where some proportion of the layers have four regions (when each layer is viewed as a ring) are determined to be chess-board vertices.  Sun et al. claim the technique works well, but note that it produces false corners from noise and is rather slow.  The scheme also relies on the thresholding producing an acceptable binary result.

There is a similar method, whose details are unpublished, included in the Parallel Tracking and Mapping for Small AR Workspaces (PTAM) reference implementation (\citealt{klein2007}).  In this the circular sampling ring from the FAST detector (\citealt{rosten2005}; \citealt{rosten2006}) is used, and upper and lower thresholds are formed which are a fixed intensity distance from the mean of the sixteen sampled points.  Proceeding around the ring the number of transitions past these thresholds are counted, and if four such transitions are found the centre point is flagged as a possible grid corner.

The results of a variety of basic detection and refinement schemes are employed in many camera calibration papers, a well known example being that of \cite{zhang2000}, but a review of such publications is beyond the scope of this work.

In general terms the Harris and Stephens detector is the one encountered most frequently in the literature, other notable papers employing it including \cite{shu2003} and \cite{douskos2007}.  Several papers such as that of \cite{lucchese2002} exist, but these detail refinement strategies to the features given by a Harris and Stephens detector.  This paper aims to offer a competing solution to the use of the Harris and Stephens detector in chess-board applications which is intrinsically more accurate (yet also amenable to the use of similar subsequent refinement strategies if the application demands it).

\section{Sampling strategy}
\label{sect:cdsampling}

When we consider an outline for an efficient chess-board corner detector, an assumption of the squares of the pattern being approximately axis-aligned with the camera's sensor would lead to a design of very low complexity, but this is obviously an excessive restriction.  A further step up the complexity scale would suggest analyzing the image for overall feature directionality and then proceed with a detector whose axes are aligned to the detected global orientation, in some respects similar to de la Escalera and Armingol's scheme, but as noted earlier, optical distortion, or use of a non-planar surface for the pattern, will lead to the grid bending significantly.  Apart from the consequence that there may then be no strong global orientation found, the wider implication is that a general purpose detector must cope with features at all orientations.  In the interest of consistency, it is obvious that such a detector must strive to award the same strength (in some sense) to features which are identical in all respects apart from orientation.

\begin{figure}
	\centring
	\begin{subfigure}{0.49\linewidth}
		\centring
		\includegraphics{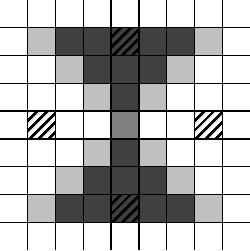}
		\caption{Four point sampling: optimal when sampling inside grid squares}
		\label{fig:r3optimal}
	\end{subfigure}
	\begin{subfigure}{0.49\linewidth}
		\centring
		\includegraphics{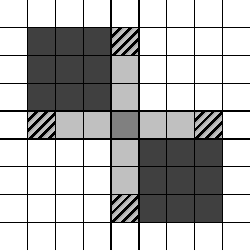}
		\caption{Four point sampling: pessimal when sampling grid square edges}
		\label{fig:r3pessimal}
	\end{subfigure}
	\begin{subfigure}{0.49\linewidth}
		\centring
		\includegraphics{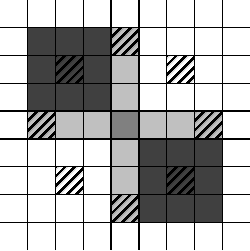}
		\caption{Eight point sampling: adequate for any vertex orientation}
		\label{fig:r3adequate}
	\end{subfigure}
	\caption{Illustration of lower bound on number of samples required to identify vertex}
	\label{fig:samp_lower_bnd}
\end{figure}

For a rotationally invariant detector, we must sample enough directions away from the centre of the feature to produce a result reliable at any angle.  In the instance of a chess-board vertex, the feature may minimally be described by finding a point where two samples taken in opposing directions from the feature centre are of one sense (say, black) and another two samples taken at a ninety degree rotation to the first pair have the opposite sense (say, white).  This gives a four point sampling pattern, as in \autoref{fig:r3optimal}, which may be viewed as a cross with intersection on the feature centre (the hatched squares denoting sampled pixels).

At this point we also consider the nature of the data commonly expected from a typical camera.  The sampled pixels around the edges of grid squares often take middling intensity values relative to the extremal intensities of pixels sampling the interiors of squares.  A couple of reasons for this phenomenon are given below.

\begin{itemize}
\item \textbf{Optical blur} due to both imperfect focus and imperfect optics.
\item \textbf{Pixel quantization} -- a single intensity value is assigned to an area of 2D optical signal, and the edges of pixels on the sensor are seldom perfectly aligned with the edges of the incoming grid image.
\end{itemize}

It is clear therefore that should the sampling cross coincide with the edges of the grid squares the result will not be reliable (\autoref{fig:r3pessimal}).  In such a case another cross in-filling the first must be used to get a result, leading to a combined sampling pattern of eight points, as illustrated in \autoref{fig:r3adequate}.  Clearly though, a vertex response in such a case is liable to, by some metric, have half the magnitude of a response where the grid edges lie between the arms of the sampling crosses and all eight samples contribute constructively.  Sampling more directions ameliorates this unevenness, as more sampling points are liable to be in an area of solid intensity rather than on a grid square edge, but at increased computational cost of processing the extra samples.

Having obtained a lower bound (of eight; we assume no FAST-like per-pixel sampling decisions) on sampling points, the issue of sample positioning becomes relevant.  In order for the feature response (however it may be calculated) at any rotation to be approximately constant, it is vital for the sampling points to be spaced at equal angles incrementally about the feature centre.  Then, considering distance away from the feature centre, two things are apparent:

\begin{enumerate}
\item Pixels close to the vertex are more likely to contain an edge than those further out, as the central angle subtended by the quasi-segment (of a pseudo-circle centred on the vertex) enclosing the pixel is much greater.  Further away from the vertex, pixels are more likely to be in areas of even intensity, in the interiors of the grid squares.  Noting the previous observations on pixels near grid square edges, sampling pixels close to the centre will lead to a weaker response.
\item Going too far away from centre risks sampling pixels from squares not forming the current feature, leading to a confused response.
\end{enumerate}

Item 1 above informs our decision to ensure distances from the centre are approximately equal -- if blur and quantization issues attenuate with distance from the feature centre it would be unfair to have certain directions sampled further out than others, violating the condition that a feature's response ought not to vary with rotation.  Approximately equal distances and equal angular spacing constrain the sampling points to be arranged in a circle, centred on the feature centre.  Furthermore, taking the two items above together, it is apparent that the radius of this circle ought to be minimized (to avoid aliasing on to other grid squares, and allow the use of more dense patterns if desired), but big enough to escape the central region of blurriness.  In some respects this circle resembles Sun et al.'s outer `layer' when using a circular window, and indeed is quite similar to that used in the PTAM code.

\begin{figure}
	\centring
	\input{Fig3}
	\caption{Similarity of r=5 circle's sampling angles}
	\label{fig:r5circle-angles2}
\end{figure}
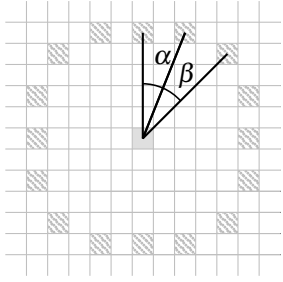

Empirically, for the majority of data considered from VGA (640 $\times$ 480 pixels) resolution cameras, a ring of radius 5 pixels (px) with sixteen samples gives a good response without constraining the minimum chess-board square size unduly, and at low computational expense.  This circle also has the desirable property that the angular sample spacing closely approximates the 22.5 degree optimal spacing of a sixteen segment circle, with sampling points spaced by either 21.8\textdegree{} or 23.2\textdegree{} (shown as $\alpha$ and $\beta$ in \autoref{fig:r5circle-angles2}).

\begin{figure}
	\centring
	\includegraphics{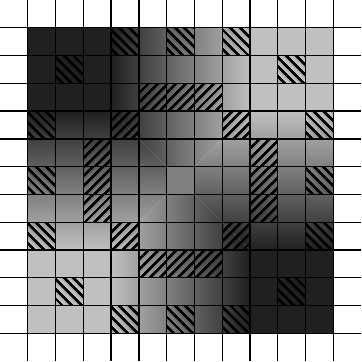}
	\caption{Smaller rings inside the blurred region contribute relatively little to improving the response}
	\label{fig:circles-blur2}
\end{figure}

The same angular spacing may be achieved with a radius 10 px circle, and use of such a circle may be appropriate in the case of highly blurred images.  The ultimate sizing of the ring is dependent on the application's optical system.  Without loss of generality a radius 5 px ring will be considered henceforth, unless stated or illustrated otherwise.

It is worth noting that employing \emph{inner} rings, as in \autoref{fig:circles-blur2}, combining a concentric radius 3 sampling circle with the radius 5 circle, is not useful, as assuming the outermost ring has been sized appropriately for the expected image blurring, any inner ring will be sampling the blurry area and have little beneficial response.  Furthermore, such techniques slow the processing of the image.  With this observation our design departs significantly from Sun et al.'s, in that it does not have multiple `layers'.

\section{Detection algorithm}

Rather than use the Sun et al./PTAM approach of performing a computationally intensive locally thresholded binarization, and then making a hard decision whether a set of samples appears to be a corner or not, some way of measuring similarity to a corner is desirable in order to provide more information to later feature consumers.  This provides an output more similar to that of the widely used Harris and Stephens detector than that of FAST.  The calculation detailed below provides this continuous quantity.

The initial grid vertex response is given by the \textit{sum response}.  When centred on a chess-board vertex, points on opposite sides of the sample circle should be of similar intensities, and the pair of points 90\textdegree{} out of phase on the circle should be of very different intensity to those at 0\textdegree{} and 180\textdegree{} phase, while being similar to each other, as previously illustrated in \autoref{fig:r3optimal}.  Taking $I_n$ as the $n^{th}$ sampling point proceeding around the sampling ring from some arbitrary starting point $I_0$, the magnitude of $(I_n + I_{n+8}) - (I_{n+4} + I_{n+12})$ should be very large when sampling around a vertex.  The \textit{sum response} (SR), so called due to the summation of opposite samples, is then given by
\begin{equation}
\mathrm{SR} = \sum_{n=0}^3 \abs{(I_n + I_{n+8}) - (I_{n+4} + I_{n+12})}
\end{equation}
and is large at a vertex point.

The most common class of false positives when using a detector simply employing the \textit{sum response} is that of those that occur along edges, though these are typically much smaller in magnitude than vertex responses.  The origin of these may be simply understood by imagining a case where one of the four sampling terms in the \textit{sum response}, say $I_{n+8}$, is one, and the rest zero.  These samples are easily seen to be consistent with an edge, and a positive response still results (though half the magnitude that would occur if $I_n$ were also one, being the vertex case).

\begin{figure}
	\centring
	\includegraphics{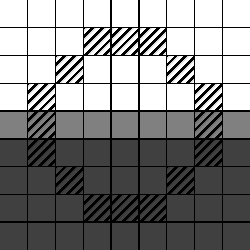}
	\caption{The case of a simple edge}
	\label{fig:simple_edge}
\end{figure}

Noting that for a simple edge such as that shown in \autoref{fig:simple_edge} (where without loss of generality a radius three circle is used for illustrative purposes), points on opposite sides of the sample circle should generally be of differing intensities; therefore the \textit{diff response} (DR), which may be expressed as
\begin{equation}
\mathrm{DR} = \sum_{n=0}^7 \abs{I_n - I_{n+8}}\text{,}
\end{equation}
should be large along edges.

Subtracting the \textit{diff response} from the \textit{sum response} forms an intermediate response with a much improved signal-to-noise ratio.  Considering the common example described above (one in four samples vastly different to the others) it may be seen that the effect of subtraction is to totally cancel the contribution of the \textit{sum response}, giving an intuitively correct intermediate response of zero.  Later we will consider how the \textit{sum} and \textit{diff response}s may be interpreted using an analogy to the DFT.

\begin{figure}
	\centring
	\begin{subfigure}{0.49\linewidth}
		\centring
		\includegraphics{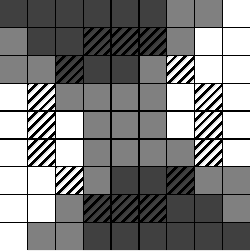}
		\caption{A corner: high response desirable}
		\label{fig:r3circle-corner}
	\end{subfigure}
	\begin{subfigure}{0.49\linewidth}
		\centring
		\includegraphics{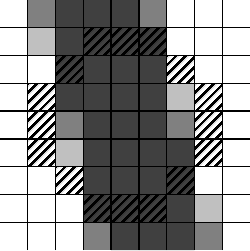}
		\caption{A stripe: rejection required}
		\label{fig:r3circle-stripe}
	\end{subfigure}
	\caption{Two very different features that have the same response on the sampling ring}
	\label{fig:stripecase}
\end{figure}

A final major false positive elimination is to remove the case where the sample circle covers a solid stripe, as in \autoref{fig:r3circle-stripe}.  Observe that the circle's samples for the corner feature shown in \autoref{fig:r3circle-corner} will be exactly the same as those for the stripe.  The two cases may only be distinguished by taking samples elsewhere -- a good location being at the centre of the ring, exploiting the aforementioned expectation of a region of intermediate intensity resulting from blur.  By computing a local intensity mean which considers a few (say, for a radius 5 px circle, 5) pixels at the centre of the sampling circle, and a larger spatial mean of all samples in the ring (\textit{neighbour mean}), an absolute difference of means, \textit{mean response}, can be found:
\begin{equation}
\mathit{mean\,response} = \abs{\mathit{neighbour\,mean} - \mathit{local\,mean}}\text{.}
\end{equation}

This will be large in the stripe case: using \autoref{fig:stripecase} as an example, the \textit{neighbour mean} will be a light grey in both cases, as will the \textit{local mean} in (a) -- leading to a small \textit{mean response}, whereas for (b) the \textit{local mean} will be much darker leading to a large absolute difference.  The \textit{mean response}, multiplied by the number of sampled pixels (16), may be subtracted from the existing response, yielding the overall response R:
\begin{equation}
\mathrm{R} = \mathit{sum\,response} - \mathit{diff\,response} - 16 \times \mathit{mean\,response}\text{.}
\end{equation}

The factor of sixteen ensures a zero overall response in the undesirable case; that where say samples $I_n$ and $I_{n+8}$ have value one and $I_{n+4}$ and $I_{n+12}$ have value zero, as does the mean of the pixels centred in the circle (i.e. the \autoref{fig:r3circle-stripe} case).
 
This overall response is not claimed to be perspective-invariant as such a claim would make no sense -- the detector does not know if the image contains perspectively distorted chess-board intersections or merely features looking like perspectively distorted chess-board intersections.  Hence a highly distorted intersection, whether distorted by perspective effects or otherwise, will still be assigned a strength, albeit one lower than that if the candidate vertex were viewed `face on'.

An additional stage to provide localized contrast/response enhancement in darker areas of the image, such as by per-pixel division by the \textit{neighbour mean}, is not employed in this detector, as the noise amplification in low intensity areas is too great.

We term this algorithm the ChESS (Chess-board Extraction by Subtraction and Summation) detector.

\subsection{DFT based interpretation of the ChESS \mbox{detector}}

If the sixteen samples taken by the sampling circle are linearized into a 1D data vector, it is seen that the FFT of this vector has a high absolute value for the second co-efficient when the circle is centred on a grid intersection, with the first co-efficient high when centred over an edge (the zeroth term being the DC term).

\begin{figure}
	\centring
	\begin{subfigure}{0.49\linewidth}
		\centring
		\includegraphics[scale=.8]{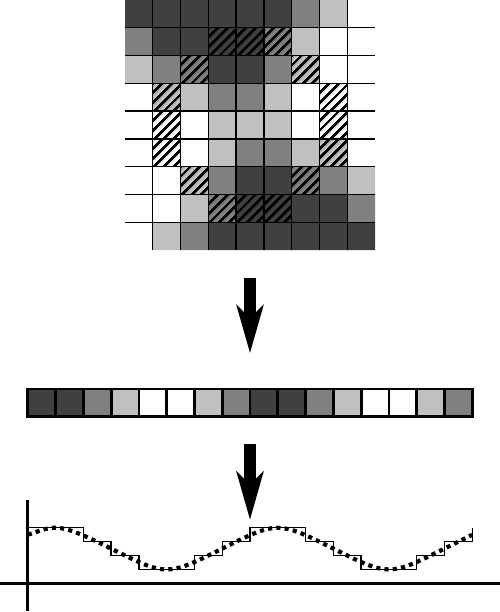}
		\caption{Second DFT co-efficient corner correlation}
		\label{fig:r3circle-corner-fft}
	\end{subfigure}
	\begin{subfigure}{0.49\linewidth}
		\centring
		\includegraphics[scale=.8]{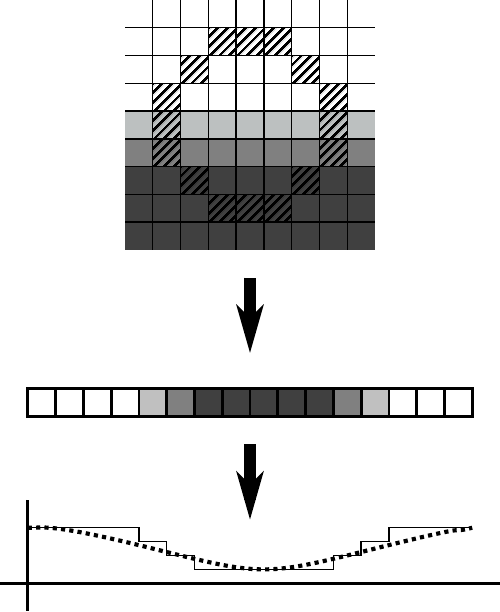}
		\caption{First DFT co-efficient edge correlation}
		\label{fig:r3circle-edge-fft}
	\end{subfigure}
	\caption{1D DFT correlation with sample circle vectors}
	\label{fig:ffts}
\end{figure}

This makes intuitive sense when considered graphically.  In \autoref{fig:r3circle-corner-fft} a corner feature is linearized clockwise from the top left sample, and the intensity values correlate well with two cycles of a cosine wave of some phase, which is akin to the DFT's second oscillatory term.  Likewise in \autoref{fig:r3circle-edge-fft} one cosine cycle is similar to the intensity vector formed from an edge.  By inspection it is apparent that any rotation of the feature described will merely result in a change of phase in the matching cosine.

It may now be seen that the \textit{sum response} attempts to perform a two cycle cosine-like match to find grid intersections, accumulating over four phases.  Similarly, the \textit{diff response} matches edges by a method not dissimilar to one matching a single cycle cosine over eight phases.

\section{Feature selection}
\label{sect:sel}

As noted in the introduction to this paper, deciding which responses are to be treated as true features is left to be determined by application specific constraints.  A few particular steps may be of use in most instances however, exploiting larger scale spatial constraints to eliminate false positive features, and these are given below.

\begin{itemize}
\item \textbf{Positive response threshold} -- discard response pixels with zero or negative intensity (since the response quantity is designed to ensure only chess-board intersections have positive intensity).
\item \textbf{Non maximum suppression} -- a standard technique to discard non-maximal responses in a small area around each pixel of the response image.  This may be used to determine integer pixel co-ordinates for a set of candidate features.
\item \textbf{Response connectivity} -- true chess-board vertex responses typically span a number of pixels; any totally isolated positive response pixels may be discarded.
\item \textbf{Neighbourhood comparison} -- comparing the magnitudes of maximal responses over a large area, those less than some proportion of the greatest responses (which are those of true chess-board features) are viewed as false features and discarded.  In many respects this compensates for the lack of intensity/contrast normalization in the detector.
\end{itemize}

\begin{figure}
	\centring
	\includegraphics[scale=6]{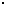}
	\includegraphics[scale=6]{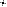}
	\includegraphics[scale=6]{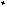}
	\includegraphics[scale=6]{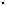}
	\includegraphics[scale=6]{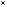}
	\caption{Responses with feature rotation of 0\textdegree{}, 11.25\textdegree{}, 22.5\textdegree{}, 33.75\textdegree{}, and 45\textdegree{}}
	\label{fig:cd5respsalign}
\end{figure}

Typical response patterns are shown for a variety of feature rotations in \autoref{fig:cd5respsalign}.  We observe that they are symmetrical about the feature centre and it can therefore be seen that for sub-pixel localization a centre of mass technique will give reasonable results, a specific example being a 5 $\times$ 5 patch centred on the maximal pixel.  This method is fast and in common use -- \cite{escalera2010} find the centre of mass of each of the points resulting from Harris detection, while \cite{sun2008} also find the centre of mass of their response clusters.  More complex refinement techniques such as those typically used to post-process features detected with the Harris and Stephens detector could also be used, but are beyond the scope of this work's aim of \emph{detecting} features initially.

\section{Orientation labelling}

The \textit{sum response} has a further use: by finding the rotation around the sampling ring at which the \textit{sum response} is maximal, each feature can be assigned one of eight orientations, relative to the pixel axes (returning to the previous DFT analogy, this is determining a quantized value for the two-cycle cosine's phase).  This labelling has many potential uses, but a trivial example is that in finding chess-board vertices the orientation labels of connected vertices ought to be in approximate anti-phase.  The details of this labelling are explained below, separately to the main detector, as while the processes could be conducted simultaneously it is often more efficient to only perform the labelling having selected a set of candidate features.

The same measure ($M$) as used in the detector's \textit{sum response} is employed; $M_n = (I_n + I_{n+8}) - (I_{n+4} + I_{n+12})$.  $\abs{M}$ gives four unique values when rotated around the 16-point sampling circle, there being eight distinct values of $M$ before duplication occurs due to symmetry, and half of those eight simply differing in sign depending on whether a given opposing pair of the four points sampled are in a ``black'' or ``white'' grid square.

To achieve the first stage of the orientation binning, for each measure an average ($AM$) across those measures one orientation either side of the current measure is found.  More explicitly, $3AM_n = M_{n-1} + M_n + M_{n+1}, \quad n \in \{ 0, 1, 2, 3 \}$ (with care taken when $n-1$ is $-1$ or $n+1$ is 4 to wrap modularly to 3 or 0 respectively and flip the sign of the $M$ in question).  The index $i$ of the orientation which has the greatest absolute average measure is taken, i.e. $i = \operatorname{arg\,max}_n \abs{AM_n}$.  To find the final orientation bin index, the sign of $M_i$ is considered, with features with positive $M_i$ being consigned to a different four orientation bins than those with a negative $M_i$.

Because a grid intersection of two black and two white squares has a rotational symmetry of order two, these eight bins correspond to increments of 22.5\textdegree{}.

With regard to chess-board decoding, this level of granularity ensures a satisfactory distance between alternate chess-board vertices, which can be viewed as 90\textdegree{} out of phase.  With the 22.5\textdegree{} distinction available here, two opposite sense features' orientations can tolerate a distance-one (in bins) orientation labelling error while remaining distinct.

\section{Experimental results}

In substantiation of the claims of robustness in detection of chess-board vertices, the ChESS detector must be compared against other detectors in use for the same problem.

\subsection{Synthetic data}
\label{ssect:synd}

Resilience to noise and invariance of response magnitude to rotation can both be quantified by simulating a feature point at varying rotations and noise levels, performing feature detection on the generated image, subsequently localizing the greatest feature response using a number of strategies, and measuring the distance of this point from the co-ordinates of the original simulated point.

Comparisons with the \cite{harris1988} algorithm and the SUSAN method (\citealt{smith1995}) are made below, the SUSAN detector being another general purpose corner detector giving a quantified feature response strength with a published reference implementation.  A further comparison is made against the PTAM detector, but this is evaluated separately due to the detector only providing a binary response.

\subsubsection{Simulation generation}
\label{sssect:simgen}

It is desirable for the simulated images to bear a reasonable resemblance to real data, in order for the results to be meaningful.  The simulation images are thus composed of four equal size rectangles in two colours, arranged to define an intersecting point.  Since a common image format of camera output is 1 channel of 8 bits the two colours are set at 64 and 191, approximately equidistant from saturation and the middle of the intensity range.

The image is then rotated by some angle around the co-ordinates of the intersection, using the ImageMagick library\footnote{\url{http://www.imagemagick.org/index.php}} with a bi-linear interpolation method specified.  The image is next cropped to VGA resolution, then a 3 $\times$ 3 Gaussian blur, using two 1D passes of a $\frac{1}{5}\begin{bmatrix}1 & 3 & 1\end{bmatrix}$ filter, roughly corresponding to a 0.675 variance, is applied.  This produces images similar to those captured by a well focussed VGA resolution camera.

Finally, noise, generated from randomly sampling a Gaussian distribution with a specified variance is added to each pixel of the image, with saturation occurring at pixel intensities of 0 and 255.

A variant on this approach may also be simulated.  The method described above may be thought of as emulating the case where the real-world edge is exactly incident on the edges of the camera's sensor elements (prior to rotation), but equally probable is the case of the edge coinciding with the middle of the elements.  By inserting a transition row and column at mid-magnitude (128) at the rectangle borders and rotating about a point offset from the centre by half a pixel in x and y this case may also be tried.  Of course, in reality the incident edge's centre will fall between these two cases, but together they ought to highlight any undesirable pathological behaviour present in these extreme cases.

\begin{figure}
	\centring
	\begin{subfigure}{0.49\linewidth}
		\centring
		\includegraphics[viewport = 23 18 53 38, clip, width=.98\linewidth]{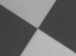}
		\caption{Captured image}
		\label{fig:realrotated}
	\end{subfigure}
	\begin{subfigure}{0.49\linewidth}
		\centring
		\includegraphics[viewport = 23 18 53 38, clip, width=0.98\linewidth]{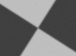}
		\caption{Simulated image}
		\label{fig:simrotated}
	\end{subfigure}
	\caption{Real and simulated images of a rotated point}
	\label{fig:rotsimreal}
\end{figure}

In \autoref{fig:simrotated} a portion of an image simulated with a rotation angle of 32.5\textdegree{} and a noise variance of 1 is shown.  \autoref{fig:realrotated} shows a similar portion of a black and white intersection captured by a real camera; the two may be seen to be visually similar.

\begin{figure}
	\centring
	\begin{subfigure}{0.49\linewidth}
		\centring
		\includegraphics[viewport = 23 18 53 38, clip, width=.98\linewidth]{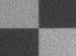}
		\caption{Captured image}
		\label{fig:realaligned}
	\end{subfigure}
	\begin{subfigure}{0.49\linewidth}
		\centring
		\includegraphics[viewport = 23 18 53 38, clip, width=0.98\linewidth]{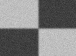}
		\caption{Simulated image}
		\label{fig:simaligned}
	\end{subfigure}
	\caption{Real and simulated noisy images of a pixel grid-aligned point}
	\label{fig:alignedsimreal}
\end{figure}

Similarly, \autoref{fig:realaligned}, an image captured with a short exposure (leading to higher noise), may be compared with \autoref{fig:simaligned}, a portion of an image simulated with no rotation, half-pixel offsetting, and a noise variance of 5, and found to again be visually similar.

\subsubsection{Effects of noise and rotation on detection}
\label{sssect:eff_of_nnr}

At a coarse level sub-pixel localization is unnecessary -- simply taking the integer pixel co-ordinates of the greatest response is sufficient to provide an overall illustration of behaviour.  A similar connectivity method to that described in \autoref{sect:sel} is employed to provide a minimal filter, discarding responses which are not connected to any of the eight adjacent pixels (horizontally, vertically and diagonally).

\begin{figure}[t]
	\centring
	\includegraphics{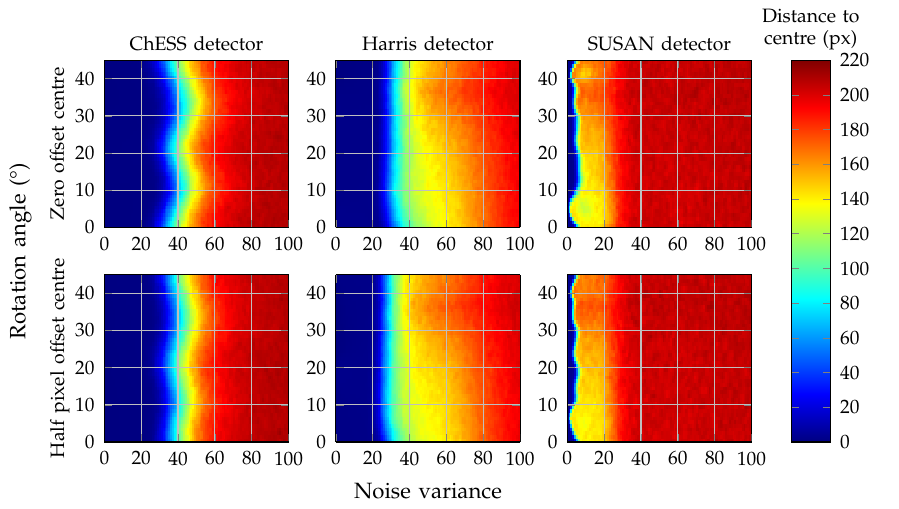}
	\caption{Basic comparison of detector performance at various feature angles and noise levels}
        \label{fig:simcomp1}
\end{figure}

\autoref{fig:simcomp1} displays the performance of the three detectors under test.  The ``ChESS detector'' is that detailed in the preceding sections; the ``Harris detector'' uses a 5 $\times$ 5 Sobel aperture, a 3 $\times$ 3 block size for the subsequent box filter, and the free parameter \textit{k} = 0.04; and the ``SUSAN detector'' uses Smith's implementation of this algorithm\footnote{Available at \url{http://users.fmrib.ox.ac.uk/~steve/susan/susan2l.c}}, with the brightness threshold value at 20.  The Harris parameters have been found empirically to give strong responses on real data, and the SUSAN threshold is the default value.  The colouring of the plots corresponds to the distance (in pixels) of the greatest response detected from where the true feature lies, i.e. the measured error increases as the colour changes from blue to red.

It can immediately be seen that the new detector performs as well or better than the Harris detector at all noise levels and rotations.  As expected the new detector's response displays periodicity about 22.5\textdegree{}, due to the angular spacing in the sampling ring discussed in \autoref{sect:cdsampling}.  It can furthermore be seen that the Harris detector's accuracy varies depending on rotation -- it is noticeably better at zero rotation than when closer to 45\textdegree{} rotation.

The SUSAN detector has not fared nearly as well as the other detectors in this test with increasing noise; the low default brightness threshold leads to noise features dominating at very low noise levels.  It may also be observed that the rotational response is uneven.

Considering the detectors in more detail, some variants must be included in the simulations for a more direct comparison.  The 5 $\times$ 5 Sobel operation applied in the Harris detector implementation has an effect of smoothing the input image by a 5 $\times$ 5 Gaussian kernel.  As the ChESS detector has no such smoothing step, it is instructive to simulate two further variants: the ChESS detector processing images smoothed by a 5 $\times$ 5 Gaussian kernel (two 1D $\frac{1}{16}\begin{bmatrix}1 & 4 & 6 & 4 & 1\end{bmatrix}$ ($\sigma^2 \approx$ 1.04) passes), and a modified Harris detector with no initial smoothing.  A further simulation of the SUSAN detector with a higher brightness threshold (40) is also warranted to ascertain its performance when detecting only strong features.

\begin{figure}
	\centring
	\includegraphics{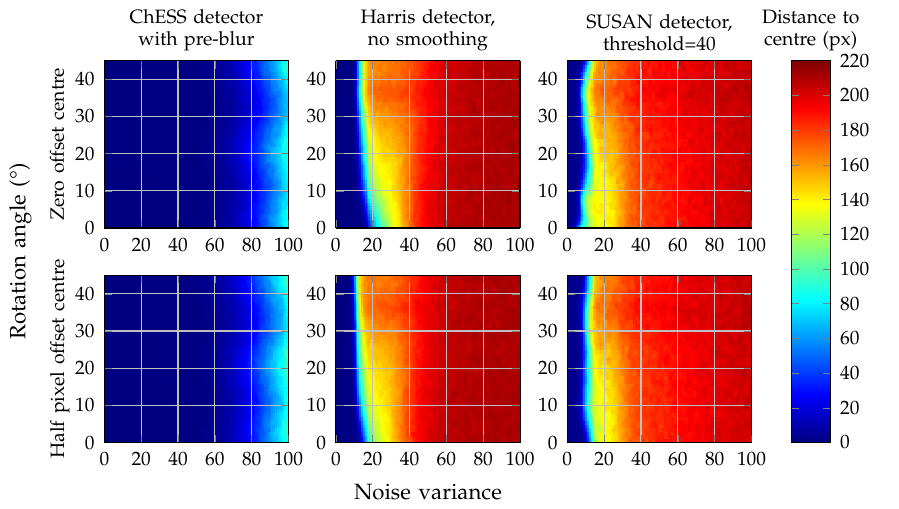}
	\caption{Further comparison of variant detector performance at various feature angles and noise levels}
        \label{fig:simcomp2}
\end{figure}

The results of simulating these variants are plotted in \autoref{fig:simcomp2}, again with the colour showing the error in terms of distance.  Blurring the input data significantly improves the new detector's resilience to noise, allowing approximately double the noise variance before significant errors occur.  Conversely, the Harris detector without the pre-blurring step becomes even more directional, and has very poor noise performance.

Setting the brightness threshold of the SUSAN detector to a larger value yields some improvement in noise resilience over the default threshold, but still does not begin to compete with the other two detectors, and in use would lead to weaker features being missed.  While the SUSAN principle is intended to not require noise reduction, we note for completeness that from simulation not presented here, using the same pre-blur as employed with the ChESS detector (and retaining the brightness threshold of 40) merely improves the noise performance to being marginally worse than the smoothed Harris detector; not a dramatic improvement.  For these reasons the SUSAN detector is not considered further in this accuracy comparison.

\begin{figure}
	\centring
	\begin{subfigure}{\linewidth}
		\centring
		\includegraphics{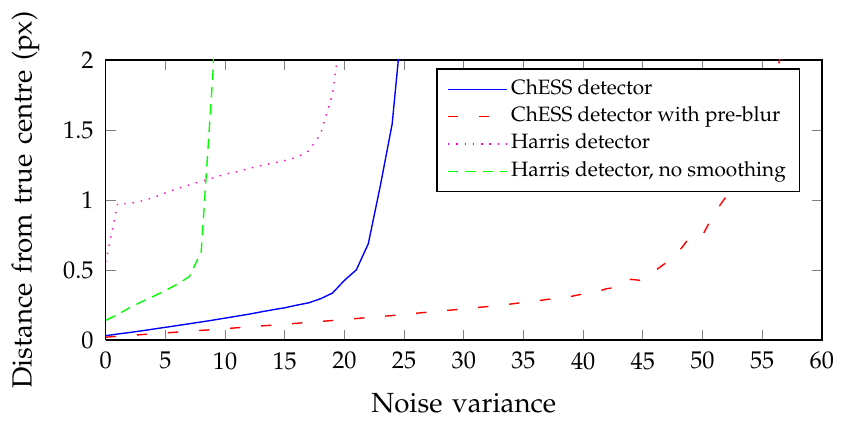}
		\caption{Performance of the detectors for varying noise (mean of all rotations)}
		\label{fig:subpixhth1}
	\end{subfigure}

	\vskip 3mm

	\begin{subfigure}{\linewidth}
		\centring
		\includegraphics{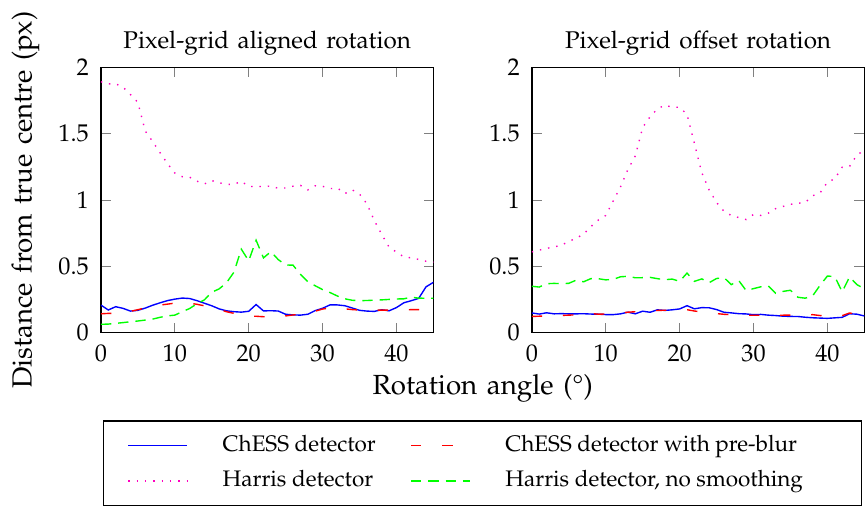}
		\caption{Performance of the detectors for varying rotation (mean of low noise regions)}
		\label{fig:subpixhth2}
	\end{subfigure}
	\caption{Error performance of the four detector schemes using 5 $\times$ 5 centre of mass sub-pixel localization method on the same axes}
	\label{fig:subpixhth}
\end{figure}

Looking in more detail at localization precision at low noise levels, \autoref{fig:subpixhth1} and \autoref{fig:subpixhth2} plot the error performance of the remaining four detector schemes, using the 5 $\times$ 5 centre of mass sub-pixel localization method described previously, on the same axes for varying noise (mean of all rotations), and rotation (mean of low noise regions) respectively.  Other localization schemes \emph{could} be employed, but it is informative to compare the ability of the raw detection method without additional complex refinement, not least to determine whether extensive post-processing is in fact necessary.

\autoref{fig:subpixhth} clearly shows that the ChESS detector variants perform better than the Harris detectors at all noise levels, and have a good and even performance at all feature rotations.  The angular performance of the two new detector variants is comparable, but the pre-blurred variant is more resilient to image noise.

\begin{figure}
	\centring
	\includegraphics{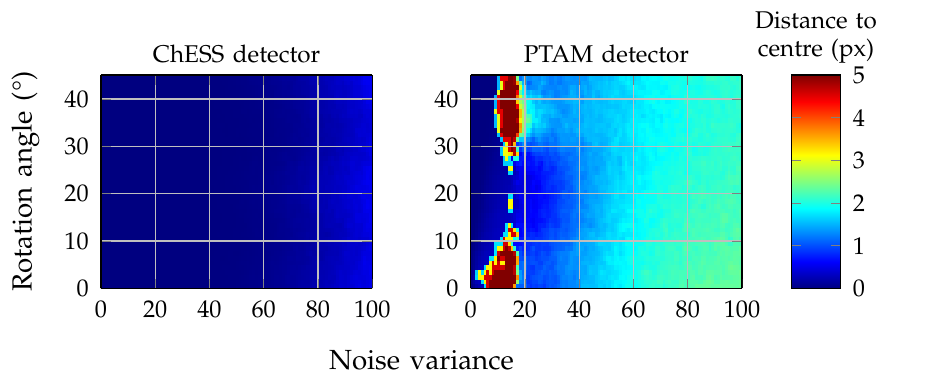}
	\caption{Basic comparison of binary response detector performance at various feature angles and noise levels}
	\label{fig:simbcomp1}
\end{figure}

To form a comparison against the PTAM detector, whose output is a per-pixel Boolean response indicating whether it is a corner feature, the output of the ChESS detector is thresholded, with the threshold set at approximately 1.5 percent of the positive response.  \autoref{fig:simbcomp1} presents the distance of the nearest detected corner feature from the true corner location, using a pixel grid aligned feature only.  The plots' colours saturate at a distance of five pixels -- any positive result detected a greater distance from the true feature is unlikely to be due to the true feature.  By default the PTAM implementation applies a Gaussian blur with $\sigma$ = 1 before sampling the image, so a Gaussian blur with similar $\sigma$ is applied prior to processing by either detector in the comparison.

It can immediately be seen that in the simulation results the PTAM detector fares worse than the ChESS detector as the noise increases.  A further poor performance region is visible under low noise conditions; this is due to the PTAM detector's rejection of corners whose central intensity is similar to the detecting region's mean, an inevitable situation with the simulated optical blur across the regions of two intensities.

\begin{figure}
	\centring
	\includegraphics{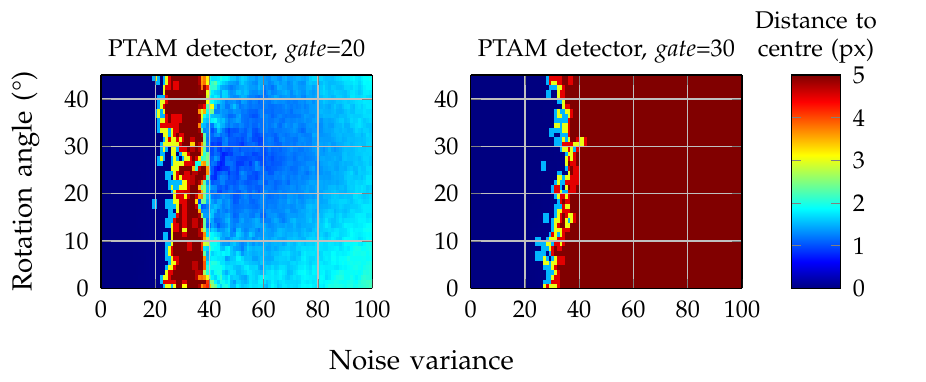}
	\caption{Further comparison of the variant PTAM detectors' performances at various feature angles and noise levels}
	\label{fig:simbcomp2}
\end{figure}

Considering only the comparatively weaker noise performance, like SUSAN the detector offers a parameter to recognize only stronger features, the \textit{gate} value.  Plots for \textit{gate}=20 and \textit{gate}=30 are shown in \autoref{fig:simbcomp2}.

Pixel intensities around the sampling ring must change by 2 $\times$ \textit{gate} in order for a white--black or black--white transition to be recorded.  Hence while the simulated noise performance of the PTAM detector is seen to improve with a higher \textit{gate} value, features would have to have black and white regions differing in intensity by more than forty (for \textit{gate}=20) in order for the feature to be recorded, whereas the new detector will still find a low contrast feature, albeit with a small response.

\subsection{Real data}
\label{ssect:reald}

While in the previous subsection attention was paid to the accuracy of the simulation, it is nevertheless true that results obtained through simulation do not always hold true in reality.  In this subsection the accuracy and robustness of the detector is validated by measuring the error and consistency in 3D reconstruction of surfaces of known shape on which a chess-board pattern is projected and multiple views of the surface recorded (a standard Structured Light technique).  The importance of accurate localization is particularly great as the extrinsic calibration of the cameras is performed on the observed data, so any error in calibration resulting from poor localization will tend to degrade the quality of the reconstruction overall.

For the camera calibration and surface reconstruction the combination of methods described in \cite{deboer2010bmvc} (and in more detail in \cite{deboer2010}) are used, drawing heavily on \cite{jl2001}.  These have been found to be reliable and accurate on a variety of data in previous studies.

The reconstruction test permits both comparison of the ChESS feature detector against other detection schemes, and testing of the ChESS detector variants against each other.  It also constructs the experiment in such a way as to test the two most likely applications for this work, namely camera calibration and 3D reconstruction.  Using a 5 $\times$ 5 centre of mass sub-pixel interpolation method in each case, the variants under test are:
\begin{itemize}
\item Harris detector (parameters as used in simulation).
\item ChESS detector without pre-blur.
\item ChESS detector with pre-blur.
\end{itemize}

The PTAM detector is not considered here due to its output not being well suited to sub-pixel feature localization.

\subsubsection{Flat plate comparison}

A flat plate is used to permit easy verification that the reconstructed surface is planar.  In the test a moving platform, whose position at any time is precisely known, moved the plate toward and away from the cameras over a travelling distance of approximately 95mm, while the distance of the plate from the cameras was around 1m.  Following the motion period the plate was held at a constant displacement; the ``rest'' period.  During the whole recorded period over 500 projected grid points were in view and these were subsequently used for calibration.

The relatively large motion is intended to result in an improved calibration of the cameras' extrinsic parameters, while the rest period permits the plate's reconstructed flatness to be compared over many frames, as no real world motion is present.

\begin{figure}
	\centring
	\includegraphics{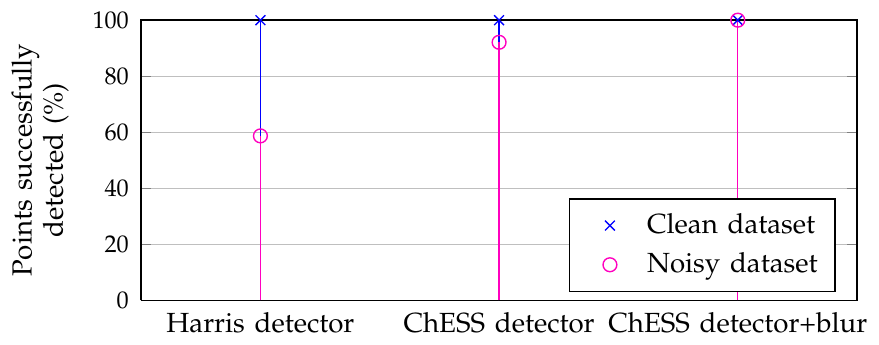}
	\caption{Performance of detectors on the flat plate data}
	\label{fig:flat_pc}
\end{figure}

The first dataset contains an optimally lit and focussed scene -- what might be considered high quality data.  \autoref{fig:flat_pc} is a plot of the percentages of grid points successfully found by each tested detection method (irrespective of their precise localization), with the results from this well-lit dataset given by the cross ($\times$) markers.  It may immediately be seen that for this ``clean'' data all the detectors are successful.

During the rest period a plane was fitted to the reconstructed surface of each frame.  The method employed was that described in \textsc{Matlab}'s documentation (\citeauthor{matlab_pca_ortho_fitting}), where a linear regression that minimizes the perpendicular distances from the data to the fitted model is found using Principle Component Analysis (PCA), forming a linear case of Total Least Squares.  This produces a unit vector normal to the plane, and summing the squared perpendicular distances from the data to the fitted plane a sum of squared errors (SSE) is calculated, indicating the quality of the fit.  Any outliers will deteriorate the quality of the fit, but since we aim for a detector that has few or no outliers it is fair to not make any special effort to exclude them separately.

\begin{figure}
	\centring
	\includegraphics[scale=.65]{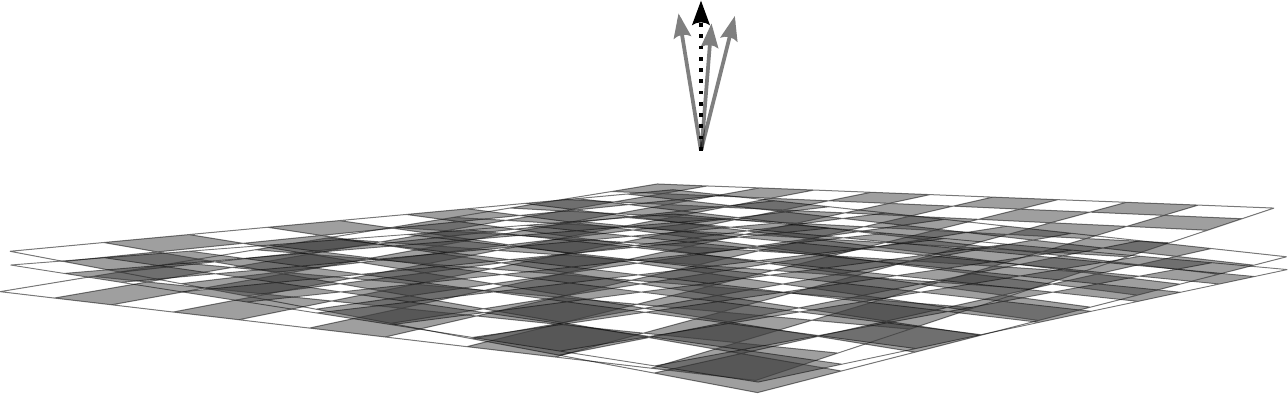}
	\caption{Illustration of various fits to a stationary flat plate, and vectors normal to the fitted planes}
	\label{fig:3d_chess_arrow}
\end{figure}

The situation is illustrated in \autoref{fig:3d_chess_arrow} where a number of planes fitted to different frames captured during the rest period are depicted, along with their normal unit vectors.  The dotted arrow shows the mean normal unit vector.

\begin{figure}
	\centring
	\includegraphics{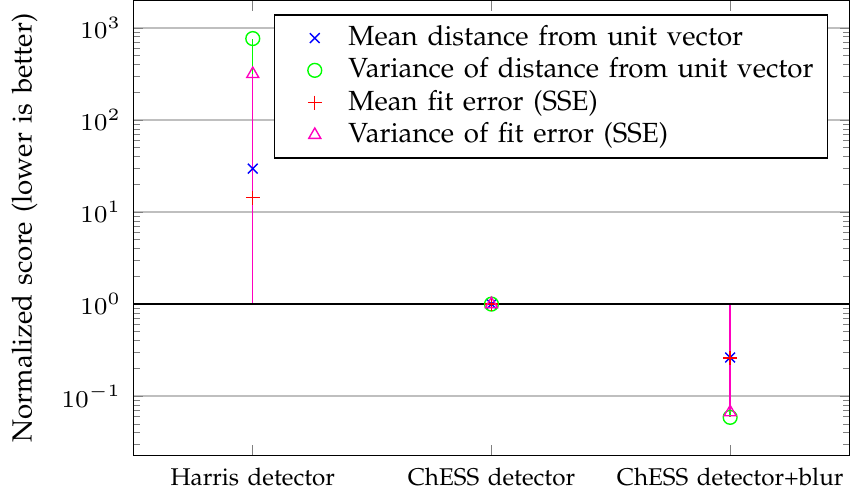}
	\caption{Comparison of statistics of the flat plate fit over the rest period for clean data, relative to those of the ChESS detector without pre-blur}
	\label{fig:clean_flat_stats}
\end{figure}

For a good stable fit, the distance of each frame's vector from this mean vector should be very small, and in \autoref{fig:clean_flat_stats} the mean and variance of these distances are plotted for all methods under test.  The mean and variance over all frames of the SSE in each frame's fit are also given.  The values are plotted relative to those of the ChESS detector without pre-blurring, which has its values normalized to one.  To provide a sense of scale, the calculated values for this normalized case have the mean distance from the mean normal unit vector (expressed as an angle due to the minute distances involved) as 0.131\textmu{}rad, with a variance of .0335\textmu{}rad$^2$, and a mean fit SSE of 6.70mm$^2$ with a variance of 0.357mm$^4$ over a patch of 100 points (per frame).  From this it can be seen that in absolute terms the error is very low -- sub-millimetre.

Compared to the detection success-rates, the fitting results show greater variety in performance between the detectors, emphasize the poorer localization resulting from the Harris detector, and reinforce the conclusion found in simulation that use of pre-blur can be beneficial when using the new detector.

\begin{figure}
	\centring
	\includegraphics{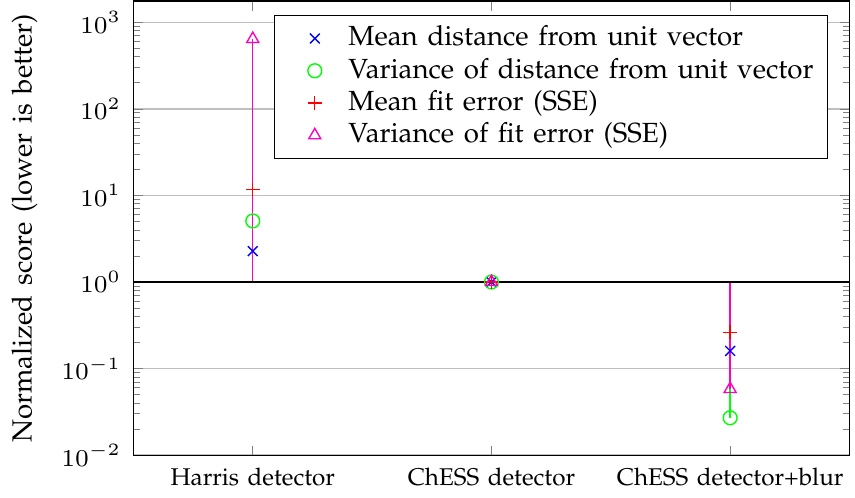}
	\caption{Comparison of statistics of the flat plate fit over the rest period for noisy data, relative to those of the ChESS detector without pre-blur}
	\label{fig:noisy_flat_stats}
\end{figure}

\begin{figure}
	\centring
	\includegraphics[scale=0.35]{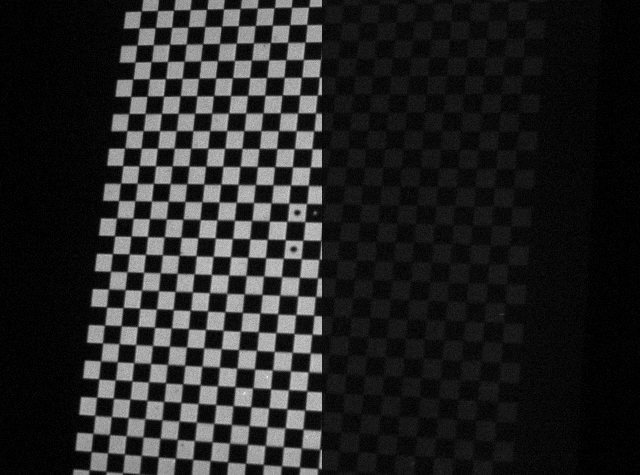}
	\caption{Visual comparison of light levels in clean and noisy captures}
	\label{fig:flat_comp}
\end{figure}

The circle markers ($\circ$) in \autoref{fig:flat_pc} and \autoref{fig:noisy_flat_stats} display the same information for a harder dataset, where the light levels are very low and hence the image noise level is much higher.  The lighting difference between the datasets may be appreciated by considering \autoref{fig:flat_comp}, though the more significant noise in the dark capture is not apparent in a still image.

The pattern of performance behaviour for the ChESS detector's variants is similar to that seen for clean data, though the improvement due to using blur is a little more pronounced.  The reference values for the ChESS detector without pre-blur have the mean distance from the mean normal unit vector as 9.13\textmu{}rad, with a variance of 181\textmu{}rad$^2$, and a mean fit SSE of 100mm$^2$ with a variance of 328mm$^4$ over a patch of 100 points (per frame).  While these figures are around an order of magnitude greater than in the clean data case, the data captured were of exceptionally poor quality.
 
\subsubsection{Cylinder comparison}

A cylinder is a more complex surface (yet reasonably easily parameterized for validation) than a flat plate, tending to distort the projected chess-board significantly due to perspective effects, and so a logical choice for a more challenging reconstruction.  For the cylinder test the same experimental procedure as for the flat plate was used, with the one change of a cylinder being the projection surface.

\begin{figure}
	\centring
	\includegraphics{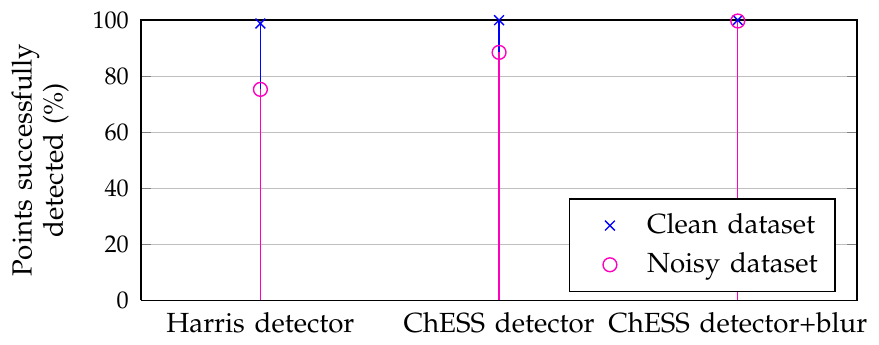}
	\caption{Performance of detectors on the cylinder data}
	\label{fig:cyl_pc}
\end{figure}

\autoref{fig:cyl_pc} shows that for low-noise data (again given by $\times$ markers) the detection is successful for all methods under test, with only the Harris detector having a less than perfect performance.

Fitting to a cylinder with arbitrary position, rotation and radius is rather harder than fitting a flat plate.  Taking the approach described in \cite{eberly2008}, a cost function for the fit to a cylinder may be expressed as
\begin{multline}
E(\mathbf{C},\mathbf{V},s) =\\\sum^n_{i=1} \left( s(\mathbf{X}_i-\mathbf{C})^T(\abs{\mathbf{V}}^2I - \mathbf{VV}^T)(\mathbf{X}_i-\mathbf{C}) - 1 \right)^2
\end{multline}
where $\mathbf{C}$ is a point on the cylinder's axis, which in turn is described by $\mathbf{V}$ (a non-unit vector, thereby allowing independent variation of its components), $s$ is related to the cylinder radius $r$ by $s = 1/(r\abs{\mathbf{V}})^2$, and $\{\mathbf{X}_i\}^n_{i=1}$ is the observed surface point set.  This permits minimization of the problem over seven parameters, and when supplied with a reasonable initial parameter set does not take many iterations to converge.

\begin{figure}
	\centring
	\includegraphics{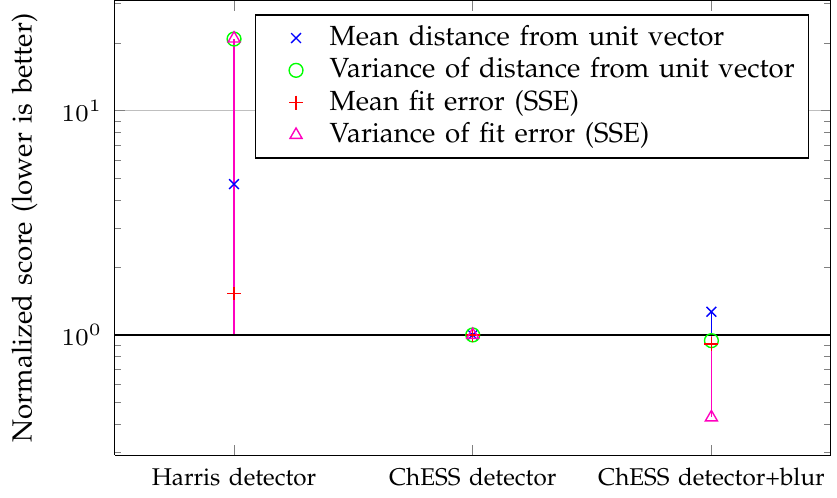}
	\caption{Comparison of statistics of the cylinder fit over the rest period for clean data, relative to those of the ChESS detector without pre-blur}
	\label{fig:clean_cyl_stats}
\end{figure}

Again, the mean and variance of the fit error across the rest period frames can be calculated, as can the mean distance from the mean axis unit vector and its variance (similarly to the method used for the flat plate's fitted normal vector).  These are plotted relative to the results of the ChESS detector without pre-blur in \autoref{fig:clean_cyl_stats} (the reference values for the axis vector distance being 0.154\textmu{}rad for the mean and 0.0476\textmu{}rad$^2$ for the variance).  The pattern of plots is much the same as for the clean flat plate data, though with the well-lit subject the empirical case for using pre-blur with the new detector is less clear.

\begin{figure}
	\centring
	\includegraphics{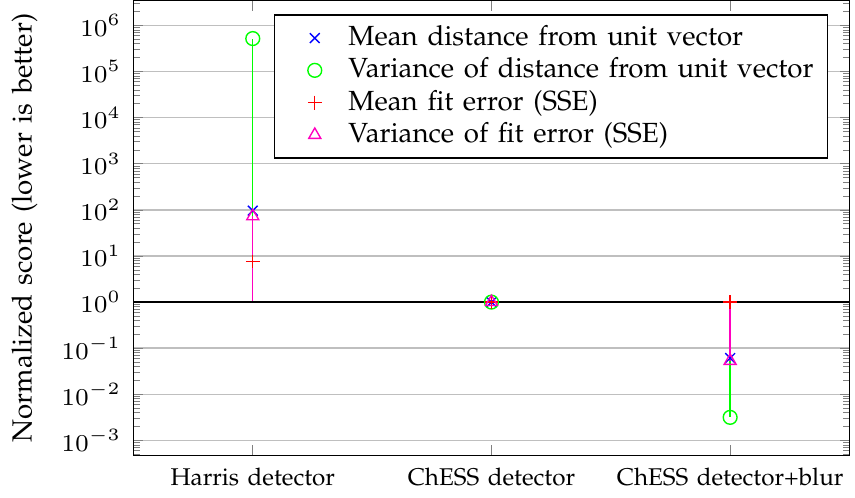}
	\caption{Comparison of statistics of the cylinder fit over the rest period for noisy data, relative to those of the ChESS detector without pre-blur}
	\label{fig:noisy_cyl_stats}
\end{figure}

\autoref{fig:cyl_pc} ($\circ$ markers) and \autoref{fig:noisy_cyl_stats} contain plots of the same measures, using the values from a noisy dataset, the reference distance values being 9.27\textmu{}rad for the mean and 184\textmu{}rad$^2$ for the variance.

It may be observed that the impact of the blurred variant is greater on noisy data.  This is in line with expectations: blur will reduce the deleterious effect of noise on the poorly lit captures.

Overall it may be seen that use of the ChESS detector allows superior reconstructions to those generated from the tested Harris detector, and use of pre-blurring can be of significant benefit on real data.

\subsection{Computational efficiency}

While accuracy and robustness are key requirements of the system, if it is to be capable of processing video in real time it is imperative that the methods used are fast.

The image processing stage of corner detection, sampling the camera image and calculating the response image, is very computationally intensive and can easily dominate the run-time of an application using its output.  A comparison of wall clock execution time to process a certain frame of VGA resolution data 5000 times on an Intel Core i5-750 processor (unless noted otherwise) is presented in this section, using three of the algorithms considered in \autoref{sssect:eff_of_nnr}: the ChESS detector, Harris and Stephens' detector, and the PTAM detector.

The detectors are all carefully implemented in the C language.  The Harris algorithm parameters are again a 5 $\times$ 5 Sobel aperture, a 3 $\times$ 3 block size for the box filter, and the free parameter \textit{k} = 0.04, and the PTAM \textit{gate}=10, both as used initially in \autoref{sssect:eff_of_nnr}.  While the algorithms do not give directly comparable output (in particular the PTAM results would require a later stage of processing to refine feature positions to sub-pixel accuracy), one could be relatively easily substituted for another in a standard tracking application.

\begin{table}[b]
	\centring
	\caption{Time spent to perform various corner detection algorithms}
	\label{tab:harris_vs_me}
	\begin{tabular}{ l | c }
		Algorithm details			& 5000 loop time (s) \\
		\hline
		ChESS detector				& 29.1 \\
		Harris and Stephens' corner detector	& 47.9 \\
		PTAM corner detector			& 39.8 \\
		\hline
		\hline
		SIMD optimized version of ChESS		& 9.6 \\
		SIMD version on Intel i7-3770 CPU	& 7.0
	\end{tabular}
\end{table}

The timing results are presented in \autoref{tab:harris_vs_me}.  While any such results will be influenced by the effort expended on code optimization, it is apparent that the new detector algorithm is highly competitive with existing approaches, taking approximately 40\% less time than Harris and Stephens' algorithm, and around 25\% less than the PTAM code.

It is important to note that the ChESS detector algorithm is well suited to further optimization using the Single Instruction, Multiple Data (SIMD) vector instructions present on most modern CPUs.  This allows the responses for multiple pixels to be processed in parallel, and the table also gives an execution time for an implementation using these instructions.  The detector is therefore capable of processing over 700 VGA resolution frames per second (fps), more than enough for real-time use in many applications.

\subsubsection{Pre-blurring}

Considering the results of \autoref{ssect:synd} and \autoref{ssect:reald}, where the benefits of pre-processing noisy data were noted, examination of the overhead of performing a 5 $\times$ 5 Gaussian blur is necessary.  Provided that a similar level of effort is expended in the implementation of the blur, the run-time addition is not onerous: a basic C language implementation adds around 15\% to the ChESS detector written in pure C, while a vectorized convolution is much more efficient and the penalty is an addition of 10\% to the SIMD detector's run-time.  In either case the burden is a relatively small hit which in situations with noisy data is clearly worthwhile.

\section{Discussion}

As demonstrated in the result section above, the fast, accurate and robust nature of the ChESS detector allows it to be employed with confidence in applications more varied than simply locating a planar or smooth chess-board patterned surface.  More varied use of chess-board patterns is common -- for example Sun et al. demonstrate pattern finding on printed non-planar sheets and patterns projected onto room corners in \cite{sun2008}.

The original motivation behind the detector's development lies in a Structured Light setting, where a chess-board pattern is projected on to a 3D object and the surface of the object reconstructed following localization of the projected grid's vertices in multiple views.  Again, chess-board patterns have been employed by others to this end, an example being in \cite{dao2010}, where Sun et al.'s method is used in the reconstruction of facial geometry.

\begin{figure}
        \centring
        \includegraphics[scale=.38]{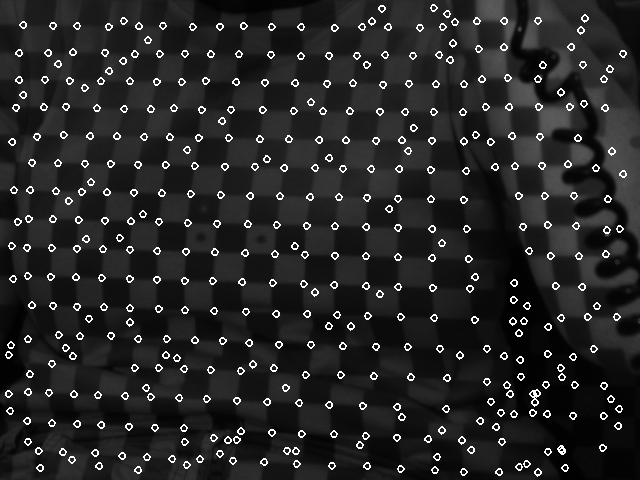}
        \caption{A sample frame of video from the Structured Light lung function measurement application, with candidate features lying under white circles}
        \label{fig:chestgrid}
\end{figure}

Our particular use of the detector is in real-time measurement of lung function in humans, observing the change in the surface of the chest of an otherwise static subject over time, as described in \cite{deboer2010bmvc}.  As the video frame in \autoref{fig:chestgrid} illustrates, detection must withstand variable lighting, poor contrast surfaces, significant perspective distortion, and potentially surface discontinuities coincident with vertices of the chess-board.  The detector presented meets these challenges routinely.

\begin{figure}
        \centring
        \includegraphics{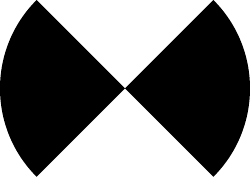}
        \caption{An optimal feature for detection}
        \label{fig:black_tie}
\end{figure}

Another avenue of work has noted that since a strong response results from any chess-board vertex-like feature, rather than necessarily requiring a chess-board pattern, a pattern of chess-board vertices will be equally detected.  This permits tiling the symbol shown in \autoref{fig:black_tie} at various rotations to form a coded grid of vertices, allowing trivial automatic correspondence determination between multiple views of the same grid.  Results using this technique are given in \cite{maldonado2011siggraph}.

\subsection{Extension to higher order intersection features}

\begin{figure}
	\setbox0=\hbox{\includegraphics[scale=1.5]{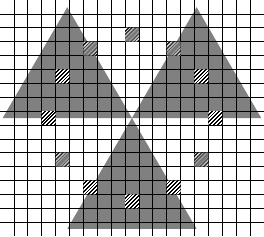}}
	\setbox1=\hbox{\includegraphics[scale=1.5]{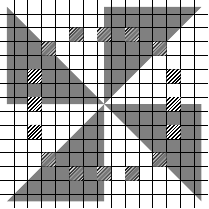}}
	\setlength{\halfdiff}{0.5\ht0-0.5\ht1}
	\centring
	\begin{subfigure}{0.49\linewidth}
		\centring
		\unhbox0
		\caption{Feature resulting from three intersecting lines}
		\label{fig:ho-3}
	\end{subfigure}
	\begin{subfigure}{0.49\linewidth}
		\centring
		\vspace{\halfdiff}
                \raisebox{\halfdiff}{\unhbox1}
		\caption{Feature resulting from four intersecting lines}
		\label{fig:ho-4}
	\end{subfigure}
	\caption{Example sampling patterns for higher order intersection features}
	\label{fig:higher_order}
\end{figure}

The matching of linearized feature neighbourhoods against arbitrary phase periodic functions, presented here in the context of chess-board pattern vertices, is applicable to higher order intersection features, though clearly at a cost of requiring higher resolution images and more sampling to maintain the level of isotropy seen in the chess-board feature detector.  Minimal sampling schemes for three and four line intersection features are illustrated in \autoref{fig:higher_order}.

The patterns in the two examples may be trivially tessellated, with the pattern in \autoref{fig:ho-3} giving a grid of identical intersections which permit unambiguous triangulation, and that in \autoref{fig:ho-4} giving an interleaved grid of intersections in both the \autoref{fig:ho-4} and chess-board styles.

Analysis of the use of such patterns remains a future avenue of work.

\section{Conclusion}

In this paper we proposed and justified the properties necessary to exclusively and uniformly detect a chess-board pattern vertex at any orientation, given common optical and sensing constraints.  From these properties we presented a simple design for a detector, which both provides a strength measure for detected features and penalizes otherwise common false positives, making its response to diverse scenes robust, all the while using relatively lightweight sampling.

Evaluation of the detector on simulated and real data has borne out its effectiveness in comparison to other freely available detectors commonly used for detection of chess-board vertices.  Particular superior function was observed in the isotropy of the response, the resilience against image noise, and in the accuracy of feature localization.

Due both to the economical sampling, and the simplicity of the operations conducted on the sampled data, the detection algorithm is very efficient and was found to be capable of a processing speed greater than other less robust and less accurate schemes considered.

The measurement of performance on real data demonstrated that while extremely well-suited to camera calibration problems, the benefits of this detector combine to permit its use in applications more varied than the detection of a planar chess-board pattern, in particular it has use in Structured Light 3D reconstruction, potentially permitting real-time processing and in detecting highly distorted chess-board patterns in general.

In the interests of others evaluating and using the ChESS detector, implementations of the algorithm are available for research purposes at \url{http://www-sigproc.eng.cam.ac.uk/~sb476/ChESS/}.

\section*{Acknowledgements}

The authors would like to thank Richard J. Wareham for the DFT analogy.

\small
\bibliographystyle{abbrvnat}
\bibliography{detector}

\end{document}

%% file: Fig3.tex
%% Creator: Inkscape inkscape 0.48.1, www.inkscape.org
%% Accompanies image file 'r5circle-angles.pdf' (pdf, eps, ps)

\begingroup
  \makeatletter
  \providecommand\color[2][]{%
    \errmessage{(Inkscape) Color is used for the text in Inkscape, but the package 'color.sty' is not loaded}
    \renewcommand\color[2][]{}%
  }
  \providecommand\transparent[1]{%
    \errmessage{(Inkscape) Transparency is used (non-zero) for the text in Inkscape, but the package 'transparent.sty' is not loaded}
    \renewcommand\transparent[1]{}%
  }
  \providecommand\rotatebox[2]{#2}
  \ifx\svgwidth\undefined
    \setlength{\unitlength}{104pt}
  \else
    \setlength{\unitlength}{\svgwidth}
  \fi
  \global\let\svgwidth\undefined
  \makeatother
  \begin{picture}(1,1.0000986)%
    \put(0,0){\includegraphics[width=\unitlength]{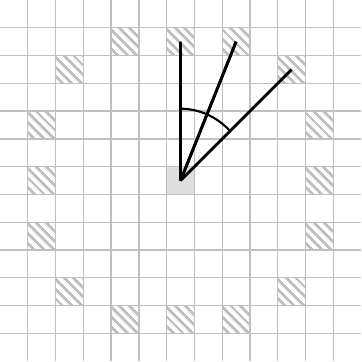}}%
    \put(0.53846154,0.76932936){\color[rgb]{0,0,0}\makebox(0,0)[lb]{\smash{$\alpha$}}}%
    \put(0.62865847,0.70568026){\color[rgb]{0,0,0}\makebox(0,0)[lb]{\smash{$\beta$}}}%
  \end{picture}%
\endgroup

%% file: detector.bbl
\begin{thebibliography}{22}
\providecommand{\natexlab}[1]{#1}
\providecommand{\url}[1]{\texttt{#1}}
\expandafter\ifx\csname urlstyle\endcsname\relax
  \providecommand{\doi}[1]{doi: #1}\else
  \providecommand{\doi}{doi: \begingroup \urlstyle{rm}\Url}\fi

\bibitem[Canny(1986)]{canny1986}
J.~F. Canny.
\newblock A computational approach to edge detection.
\newblock \emph{{IEEE} Transactions on Pattern Analysis and Machine
  Intelligence}, 8:\penalty0 679--698, November 1986.
\newblock ISSN 0162-8828.
\newblock \doi{10.1109/TPAMI.1986.4767851}.
\newblock URL \url{http://portal.acm.org/citation.cfm?id=11274.11275}.

\bibitem[Dao and Sugimoto(2010)]{dao2010}
V.~N. Dao and M.~Sugimoto.
\newblock A robust recognition technique for dense checkerboard patterns.
\newblock In \emph{2010 20th International Conference on Pattern Recognition},
  ICPR 2010, pages 3081--3084, Washington, DC, USA, Aug. 2010. IEEE Computer
  Society.
\newblock ISBN 978-0-7695-4109-9.
\newblock \doi{10.1109/ICPR.2010.755}.

\bibitem[de~Boer et~al.(2010)de~Boer, Lasenby, Cameron, Wareham, Ahmad, Roach,
  Hills, and Iles]{deboer2010bmvc}
W.~de~Boer, J.~Lasenby, J.~Cameron, R.~Wareham, S.~Ahmad, C.~Roach, W.~Hills,
  and R.~Iles.
\newblock {SLP}: A zero-contact non-invasive method for pulmonary function
  testing.
\newblock In \emph{Proceedings of the British Machine Vision Conference}, pages
  85.1--85.12. BMVA Press, 2010.
\newblock ISBN 1-901725-40-5.
\newblock \doi{doi:10.5244/C.24.85}.
\newblock URL
  \url{http://www.bmva.org/bmvc/2010/conference/paper85/paper85.pdf}.

\bibitem[de~Boer(2010)]{deboer2010}
W.~H. de~Boer.
\newblock {S}tructured {L}ight {P}lethysmography: A non-invasive method for
  pulmonary function testing using visible light.
\newblock Master's thesis, Signal Processing and Communications Group,
  Cambridge University Engineering Department, August 2010.

\bibitem[de~la Escalera and Armingol(2010)]{escalera2010}
A.~de~la Escalera and J.~M. Armingol.
\newblock Automatic chessboard detection for intrinsic and extrinsic camera
  parameter calibration.
\newblock \emph{Sensors}, 10\penalty0 (3):\penalty0 2027--2044, 2010.
\newblock ISSN 1424-8220.
\newblock \doi{10.3390/s100302027}.
\newblock URL \url{http://www.mdpi.com/1424-8220/10/3/2027/}.

\bibitem[Douskos et~al.(2007)Douskos, Kalisperakis, and Karras]{douskos2007}
V.~Douskos, I.~Kalisperakis, and G.~Karras.
\newblock Automatic calibration of digital cameras using planar chess-board
  patterns.
\newblock In \emph{Proceedings of the 8th Conference on Optical {3D}
  Measurement Techniques}, volume~1, pages 132--140. Wichman Verlag, 2007.
\newblock URL
  \url{http://citeseerx.ist.psu.edu/viewdoc/summary?doi=10.1.1.99.5535}.

\bibitem[Eberly(2008)]{eberly2008}
D.~Eberly.
\newblock Fitting {3D} data with a cylinder, February 2008.
\newblock URL
  \url{http://www.geometrictools.com/Documentation/CylinderFitting.pdf}.

\bibitem[Haralick et~al.(1987)Haralick, Sternberg, and Zhuang]{haralick1987}
R.~M. Haralick, S.~R. Sternberg, and X.~Zhuang.
\newblock Image analysis using mathematical morphology.
\newblock \emph{{IEEE} Transactions on Pattern Analysis and Machine
  Intelligence}, 9:\penalty0 532--550, July 1987.
\newblock ISSN 0162-8828.
\newblock \doi{10.1109/TPAMI.1987.4767941}.

\bibitem[Harris and Stephens(1988)]{harris1988}
C.~Harris and M.~Stephens.
\newblock A combined corner and edge detector.
\newblock In \emph{Proceedings of The Fourth Alvey Vision Conference},
  volume~15, pages 147--151. Manchester, UK, 1988.
\newblock URL
  \url{http://www.assembla.com/spaces/robotics/documents/abzMnAOEer3zB7ab7jnrAJ/download/harris88.pdf}.

\bibitem[Klein and Murray(2007)]{klein2007}
G.~Klein and D.~Murray.
\newblock Parallel tracking and mapping for small {AR} workspaces.
\newblock In \emph{Proceedings of the Sixth {IEEE} and {ACM} International
  Symposium on Mixed and Augmented Reality}, ISMAR '07, pages 1--10,
  Washington, DC, USA, November 2007. IEEE Computer Society.
\newblock ISBN 978-1-4244-1749-0.
\newblock \doi{10.1109/ISMAR.2007.4538852}.
\newblock URL
  \url{http://www.robots.ox.ac.uk/~gk/publications/KleinMurray2007ISMAR.pdf}.

\bibitem[Lasenby and Stevenson(2001)]{jl2001}
J.~Lasenby and A.~X.~S. Stevenson.
\newblock Using geometric algebra for optical motion capture.
\newblock In E.~B. Corrochano and G.~Sobczyk, editors, \emph{Geometric Algebra
  with Applications in Science and Engineering}, pages 147--167.
  Birkh\"{a}user, Boston, MA, USA, 2001.
\newblock ISBN 0817641998.

\bibitem[Lucchese and Mitra(2002)]{lucchese2002}
L.~Lucchese and S.~K. Mitra.
\newblock Using saddle points for subpixel feature detection in camera
  calibration targets.
\newblock In \emph{2002 Asia-Pacific Conference on Circuits and Systems},
  volume~2 of \emph{APCCAS '02}, pages 191 -- 195 vol.2, 2002.
\newblock \doi{10.1109/APCCAS.2002.1115151}.

\bibitem[Maldonado and Lasenby(2011)]{maldonado2011siggraph}
T.~J. Maldonado and J.~Lasenby.
\newblock Simulation of breathing for medical applications.
\newblock In \emph{ACM SIGGRAPH 2011 Posters}, SIGGRAPH '11, pages 7:1--7:1,
  New York, NY, USA, 2011. ACM.
\newblock ISBN 978-1-4503-0971-4.
\newblock \doi{http://doi.acm.org/10.1145/2037715.2037723}.
\newblock URL \url{http://doi.acm.org/10.1145/2037715.2037723}.

\bibitem[Rosten and Drummond(2005)]{rosten2005}
E.~Rosten and T.~Drummond.
\newblock Fusing points and lines for high performance tracking.
\newblock In \emph{Proceedings of the Tenth IEEE International Conference on
  Computer Vision}, volume~2 of \emph{ICCV 2005}, pages 1508--1511, October
  2005.
\newblock \doi{10.1109/ICCV.2005.104}.
\newblock URL
  \url{http://mi.eng.cam.ac.uk/~er258/work/rosten_2005_tracking.pdf}.

\bibitem[Rosten and Drummond(2006)]{rosten2006}
E.~Rosten and T.~Drummond.
\newblock Machine learning for high-speed corner detection.
\newblock In \emph{Proceedings of the 9th European Conference on Computer
  Vision}, volume~1 of \emph{ECCV 2006}, pages 430--443, May 2006.
\newblock \doi{10.1007/11744023_34}.
\newblock URL
  \url{http://mi.eng.cam.ac.uk/~er258/work/rosten_2006_machine.pdf}.

\bibitem[Shu et~al.(2003)Shu, Brunton, and Fiala]{shu2003}
C.~Shu, A.~Brunton, and M.~Fiala.
\newblock Automatic grid finding in calibration patterns using {D}elaunay
  triangulation.
\newblock Technical Report NRC-46497/ERB-1104, National Research Council
  Canada, Aug. 2003.
\newblock URL
  \url{http://www.scs.carleton.ca/~c_shu/Research/Projects/CAMcal/gridfind_report.pdf}.

\bibitem[Smith and Brady(1997)]{smith1995}
S.~M. Smith and J.~M. Brady.
\newblock {SUSAN}--{A} new approach to low level image processing.
\newblock \emph{International Journal of Computer Vision}, 23:\penalty0 45--78,
  May 1997.
\newblock ISSN 0920-5691.
\newblock \doi{10.1023/A:1007963824710}.
\newblock URL \url{http://portal.acm.org/citation.cfm?id=258049.258056}.

\bibitem[Soh et~al.(1997)Soh, Matas, and Kittler]{soh1997}
L.~Soh, J.~Matas, and J.~Kittler.
\newblock Robust recognition of calibration charts.
\newblock In \emph{Sixth International Conference on Image Processing and its
  Applications, 1997}, volume~2 of \emph{IPA '97}, pages 487 --491. IEE, July
  1997.
\newblock \doi{10.1049/cp:19970941}.

\bibitem[Sun et~al.(2008)Sun, Yang, Xiao, and Hu]{sun2008}
W.~Sun, X.~Yang, S.~Xiao, and W.~Hu.
\newblock Robust checkerboard recognition for efficient nonplanar geometry
  registration in projector-camera systems.
\newblock In \emph{Proceedings of the 5th ACM/IEEE International Workshop on
  Projector Camera Systems}, PROCAMS '08, pages 1--7, New York, NY, USA, 2008.
  ACM.
\newblock ISBN 978-1-60558-272-6.
\newblock \doi{10.1145/1394622.1394625}.

\bibitem[{The MathWorks Inc.}()]{matlab_pca_ortho_fitting}
{The MathWorks Inc.}
\newblock Statistics toolbox -- fitting an orthogonal regression using
  principal components analysis.
\newblock URL
  \url{http://www.mathworks.com/products/statistics/demos.html?file=/products/demos/shipping/stats/orthoregdemo.html}.

\bibitem[Yu and Peng(2006)]{yu2006}
C.~Yu and Q.~Peng.
\newblock Robust recognition of checkerboard pattern for camera calibration.
\newblock \emph{Optical Engineering}, 45\penalty0 (9):\penalty0 093201, 2006.
\newblock \doi{10.1117/1.2352738}.
\newblock URL \url{http://link.aip.org/link/?JOE/45/093201/1}.

\bibitem[Zhang(2000)]{zhang2000}
Z.~Zhang.
\newblock A flexible new technique for camera calibration.
\newblock \emph{{IEEE} Transactions on Pattern Analysis and Machine
  Intelligence}, 22\penalty0 (11):\penalty0 1330--1334, Nov. 2000.
\newblock ISSN 0162-8828.
\newblock \doi{10.1109/34.888718}.
\newblock URL \url{http://dx.doi.org/10.1109/34.888718}.

\end{thebibliography}
